# The Impact of Example Selection in Few-shot Prompting on Automated Essay Scoring Using GPT Models

Lui Yoshida

The University of Tokyo, Tokyo, Japan
luiyoshida@g.ecc.u-tokyo.ac.jp

**Abstract.** This study investigates the impact of example selection on the performance of automated essay scoring (AES) using few-shot prompting with GPT models. We evaluate the effects of the choice and order of examples in few-shot prompting on several versions of GPT-3.5 and GPT-4 models. Our experiments involve 119 prompts with different examples, and we calculate the quadratic weighted kappa (QWK) to measure the agreement between GPT and human rater scores. Regression analysis is used to quantitatively assess biases introduced by example selection. The results show that the impact of example selection on QWK varies across models, with GPT-3.5 being more influenced by examples than GPT-4. We also find evidence of majority label bias, which is a tendency to favor the majority label among the examples, and recency bias, which is a tendency to favor the label of the most recent example, in GPT-generated essay scores and QWK, with these biases being more pronounced in GPT-3.5. Notably, careful example selection enables GPT-3.5 models to outperform some GPT-4 models. However, among the GPT models, the June 2023 version of GPT-4, which is not the latest model, exhibits the highest stability and performance. Our findings provide insights into the importance of example selection in few-shot prompting for AES, especially in GPT-3.5 models, and highlight the need for individual performance evaluations of each model, even for minor versions.

## 1 Introduction

Large Language Models (LLMs) can perform tasks in various domains through user interaction [1, 2] and are expected to impact education significantly, and their potential use in education has already been explored [3, 4, 5]. One promising application is automated essay scoring (AES), and initial studies have begun to investigate this area [6-8]. For instance, Yancey et al. [7] validated the performance of GPT-3.5 and GPT-4 on AES using their own dataset and confirmed that including scoring examples in the prompt improves performance. Notably, GPT-4 outperformed GPT-3.5 and achieved expert-level performance.

Several issues need to be addressed when using LLMs for AES. Prompt design is crucial when utilizing LLMs, and few-shot learning, which includes examples of output, is a powerful method for improving general performance [9, 10]. However, research in sentiment analysis [11] has reported that the choice of examples can bias the results. Nevertheless, previous studies on AES have not provided insights into the selection of examples. Therefore, there is a possibility that performance could be further improved by carefully selecting examples, but this has not been verified.

Furthermore, investigating different versions of LLMs is also important. If the performance of GPT-3.5, which is approximately 1/10th the cost of GPT-4, can be improved by selecting appropriate examples, it would be useful from an educational practice perspective. However, this has not been clarified. Moreover, it is unclear whether the impact of example selection is consistent across models, including minor versions. If it is consistent, the insights obtained from one model can be applied to other models; otherwise, each model will need to be validated individually in the future.

Therefore, in this study, we evaluate the effects of the choice and order of examples in few-shot prompting, using several GPT models. First, we perform AES using 119 prompts with different examples and calculate quadratic weighted kappa (QWK), the degree of agreement between GPT and human rater scores. Then, we use regression analysis to assess biases of example selection quantitatively.

## 2   Background

**LLMs and Prompting.** Since the output of LLMs is highly dependent on the prompt, many prompting techniques have been developed. For example, few-shot prompting, which includes multiple examples in the prompt, enabled the best performance in various tasks at the time [9, 10]. Research has also demonstrated that employing a Chain of Thought (CoT) prompting, which describes the process of solving a task, can enhance performance [12, 13]. Furthermore, even zero-shot prompts that do not provide examples can improve the performance of LLMs; for example, simply adding "Let's think step by step" can realize CoT and significantly improve task execution performance [14, 15].

Few-shot prompting, which is the focus of this study, is important because it significantly impacts performance. However, bias has been reported where the choice and order of the examples presented affect the results [11, 16, 17]. For example, Zhao et al. [11] pointed out that the accuracy of few-shot prompting is unstable and that the order and type of concrete examples can cause bias and significantly change the results. Specifically, in sentiment analysis, it was confirmed that there is a majority label bias, where the more labels (negative or positive) given to examples, the more the results are given that label, and a recency bias, where the label of the last example given appears more frequently in the results. However, it is unclear whether these representative biases exist in AES, and it is also unknown whether countermeasures are necessary.

**AES.** AES has been studied for more than 50 years, beginning with Project Essay Grade by Page [18]. In the initial paradigm, researchers manually designed features and created labeled data which were utilized in multiple regression analysis, Latent Semantic Analysis, and machine learning [18-20] in 1960s to 2000s. From the 2000s, neural networks, which do not require manual feature design, began to emerge, shifting the next paradigm. The focus of design moved towards network architecture, including layer structures and propagation mechanisms [21-23]. Then, the introduction of the Transformer [24] marked the beginning of the third paradigm, with the development of models such as GPT [9] and Bidirectional Encoder Representations from Transformers (BERT) [25], employing a pre-training and fine-tuning approach. Particularly, BERT has shown high performance in AES, achieving state-of-the-art results with methods from this paradigm [26-28].

The emergence of LLMs may initiate a new paradigm in AES. By utilizing LLMs in AES, we could overcome the challenges of conventional AES, such as the lack of evaluation based on the content and domain knowledge and the unclear evaluation criteria in neural network-based AES [19, 29]. However, only a few studies have been accumulated [6-8]. For example, Yancey et al. [7] performed AES using GPT-3.5 and GPT-4 with prompt engineering techniques. The results confirmed that the performance was higher when examples were included in a prompt and showed that the evaluation by GPT-4 was almost as good as the expert level. However, there are few studies on selecting and ordering the examples, and there is no knowledge of bias by examples in AES using LLMs.

Through our study, we could be able to gain insights that can further improve performance. Furthermore, by evaluating the impact on different versions of LLMs, we could obtain practically useful knowledge, such as whether there are commonalities between models and whether lower versions can also achieve high performance.

## 3 Methods

### 3.1 Automated Essay Scoring Using GPT Models

**Dataset.** We used TOEFL11 [30] as the essay dataset. The dataset contains eight essay prompts and their corresponding examinee essays, each with approximately 1,000 to 1,600 essays, for a total of 12,100 essays. The data also includes expert ratings of the essays on a three-point scale of high, medium, and low. These ratings were first evaluated by several experts using a 5-point rubric and finally compressed to a 3-point rating according to a set of rules. The rubric ratings are not included in the dataset.

The number of essay scoring by the GPT models in this study is the number of models multiplied by the number of prompts and the number of essays to be evaluated. Therefore, the number of essay scoring is large when all the essays are used; then we sampled the essay data to be evaluated by GPT models. We selected three essays each from the eight essay prompts with a rating value of high, medium, and low, respectively, for a total of 72 essays for evaluation.

**GPT Models.** In this study, to evaluate the impact of example selection across models, we utilized three instances each of OpenAI's GPT-3.5 and GPT-4 models. For GPT-3.5, we used models released in June 2023 (gpt-3.5-turbo-0613 denoted as Jun23), November 2023 (gpt-3.5-turbo-1106 denoted as Nov23), and January 2024 (gpt-3.5-turbo-0125 denoted as Jan24). Regarding GPT-4, we employed models released in June 2023 (gpt-4-0613 denoted as Jun23), November 2023 (gpt-4-1106-preview denoted as Nov23), and January 2024 (gpt-4-0125-preview denoted as Jan24). The system prompt was not used. The following prompts were used to obtain the API response. For parameters, temperature was set to 0, and default values were used for all other parameters.

**Bias.** In this study, we treat essay scores (high, medium, and low) of examples in few-shot prompting as labels. We define majority label bias as the bias introduced in essay evaluation by LLMs due to the number of essays exemplifying a specific score. Additionally, we define recency bias as the bias in essay evaluation attributed to the score of the last exemplified essay.

**Prompts.** We developed prompts based on those used by Yancey et al. [7]. The prompts comprised several components: Instruction, Essay Prompt, Response, Rubric, Rating Examples, and Output Format (see **Fig. 1**). The Instruction section described the essay evaluation task, while the Essay Prompt section presented the prompt for the essay. The Response section contained the essay to be evaluated. Lastly, the Output Format section was the template for output.

We prepared prompts categorized into four main groups. The first category involves zero-shot prompts that do not include rating examples, labeled as category N. The second through fourth categories consist of few-shot prompts, which include one to three rating examples. Given that evaluation scores are categorized into three levels, 1-shot, 2-shot, and 3-shot prompts have 3, 9, and 27 variants, respectively. Each category is named by concatenating the capital initials (H: high, M: medium, L: low) of the evaluation scores of the examples in order. For instance, a 1-shot prompt with a high evaluation score is categorized as "H", a 2-shot prompt with evaluation scores in the order of medium and low as "ML", and a 3-shot prompt with scores in the order of low, medium, and high as "LMH". While the rubric is based on a 5-point scale, the dataset only includes three types of scores; therefore, in our experiments, the evaluation scores for examples mentioned in few-shot prompts are substituted with high, medium, and low as 4, 3, and 2, respectively. Prior to this experiment, 200 essays were evaluated with GPT-3.5 (Jun23) using three types of 3-shot prompts with the values for high, medium, and low altered to [4.5, 3, 1.5], [4, 3, 2], and [5, 3, 1], respectively. The

---

You are a rater for writing responses on a high-stakes English language exam for second language learners. You will be provided with a prompt and the test-taker's response. Your rating should be based on the rubric below, following the specified format. There are rating samples of experts so that you can refer to those when rating.

\# Prompt
"""*Essay prompt*"""

\# Response
"""*Essay to be evaluated*"""

\# Rubric
*Rubric*

\# Rating samples of experts
*Examples*

\# Output format:
Rating: [<<<Your rating here.>>>]

**Fig. 1.** A template of a prompt. The parts where data should be inserted are in *italics*.

average QWK was highest for the values [4, 3, 2], which led to these values for this experiment.

When creating prompts for each category, we prepared three sets of examples to ensure they did not overlap with the essay data being evaluated. For instance, for the HHH category prompts, we extracted three essays with high evaluation scores three times randomly to prepare three prompts for the HHH category. We prepared a total of 117 few-shot prompts, which is the sum of the categories for 1, 2, and 3-shot prompts (39) multiplied by three sets. Since the 0-shot prompt does not include examples, there was only one prompt, making a total of 118 prompts prepared across all categories.

**GPT Ratings.** We obtained 50,976 GPT ratings by using the API of six GPT models described in **Models** during January and February 2024, targeting 72 essays with 118 prepared prompts. Two responses within the category LM by GPT-3.5 (Nov23) did not yield ratings; therefore, these instances were excluded from our analysis.

**Agreement between Experts and GPT Models.** To evaluate the agreement between experts and GPT models, we used the QWK, a widely used metric to assess the concordance between machine and human evaluations [31, 32]. Since GPT evaluations used a rubric out of 5 points, we converted these into three levels: scores above 3 as high, 3 as medium, and below 3 as low, and numerically coded these as 3, 2, 1, respectively, to calculate the QWK.

### 3.2 Regression Analysis for Bias Evaluation

To assess the impact of majority bias and recency bias from few-shot prompt examples on AES using GPT models, we conducted two regression analyses. One analyzed the influence on GPT-generated essay scores, and the other evaluated the effect on QWK. After outlining the common elements of both analyses, we describe each in detail.

**Common Component.** Both analyses shared independent variables to assess the impact of majority label bias, using the number of examples rated as high, medium, or low (denoted as $H_n, M_n, L_n$), and to evaluate the effect of recency bias, employing variables indicating if the last example was high, medium, or low (1) or not (0) (denoted as $H_l, M_l, L_l$). To compare regression coefficients between models in each analysis, these independent variables were standardized to have a mean of 0 and a variance of 1. Due to the small number of examples in 1-shot and 2-shot prompts, only the results of the 3-shot prompts were used for this analysis.

**Bias Influence on Scores.** The dependent variable for the regression analysis on essay scores was the GPT-generated essay score ($Score$), which was standardized by dividing by the maximum rating value of 5. The regression model can be described as follows, where the coefficients of each independent variable are denoted by $\beta^s_{H_n}, \beta^s_{M_n}, \beta^s_{L_n}, \beta^s_{H_l}, \beta^s_{M_l}, \beta^s_{L_l}$, and the intercept by $\alpha_s$:

$$Score = \beta^s_{H_n} H_n + \beta^s_{M_n} M_n + \beta^s_{L_n} L_n + \beta^s_{H_l} H_l + \beta^s_{M_l} M_l + \beta^s_{L_l} L_l + \alpha_s \quad (1)$$

For each essay, regression analysis was performed for each GPT model, calculating the regression coefficients and the coefficient of determination ($R^2$). A meta-analysis, typically suitable for integrating multiple regression analysis results by treating regression coefficients as effect sizes, was not possible due to essays where all coefficients were zero. Therefore, to test for the presence of bias, i.e., whether the estimated regression coefficients were non-zero, a one-sample t-test at the significance level of 0.05 was conducted, with Holm's correction applied for multiple comparisons.

**Bias Influence on QWK.** The dependent variable for the regression analysis on QWK was QWK itself. This regression model can also be described using a similar formula where the coefficients of each independent variable are denoted by $\beta_{H_n}^q, \beta_{M_n}^q, \beta_{L_n}^q, \beta_{H_l}^q, \beta_{M_l}^q, \beta_{L_l}^q$, and the intercept by $\alpha_q$:

$$QWK = \beta_{H_n}^q H_n + \beta_{M_n}^q M_n + \beta_{L_n}^q L_n + \beta_{H_l}^q H_l + \beta_{M_l}^q M_l + \beta_{L_l}^q L_l + \alpha_q \qquad (2)$$

Regression analysis was performed, calculating the regression coefficients and the coefficient of determination ($R^2$) for each model. A t-test at the significance level of 0.05 was used to assess whether the estimated regression coefficients were non-zero, with Holm's correction for multiple comparisons.

## 4  Results

### 4.1  QWK between Experts and GPT Models

The average and standard deviation of the QWK across each category of all GPT models are shown in **Fig. 2**. The model with the highest average QWK across all categories was GPT-4 (Jun23), while the lowest was GPT-3.5 (Jun23). It was observed that the GPT-3.5 models were more influenced by the example selection compared to the GPT-4 models.

To examine the trends in evaluations for each model, the distribution of GPT evaluations for zero-shot (N), the category with the lowest average QWK, and the highest average QWK are presented in **Fig. 3**. Given that the number of expert evaluations for each score level is equal in our data, the ideal scenario is for the number of each score level obtained by GPT to be equal as well. For the GPT-3.5 models, results in N generally showed a tendency to overestimate scores in Jun23, underestimate ones in Nov23, and estimate more scores as medium in Jan24. For the GPT-4 models, Jun23 showed a relatively unbiased evaluation, while Nov23 and Jan24 tended to underestimate scores.

### 4.2  Regression Analysis for Bias Evaluation

The results of the regression analysis on essay scores for each model are documented in **Table 1**. Although the coefficients of determination were generally low, significant recency bias and majority label bias were identified in the GPT-3.5 models. In the GPT-4 models, some significant recency bias was observed, but majority label bias was not found. Comparison of the absolute values of the regression coefficients showed that recency bias had a larger impact than majority label bias. Furthermore, the absolute values of significant regression coefficients were larger for GPT-3.5 than for GPT-4, indicating that the influence of bias from examples on essay scores was greater for GPT-3.5 than for GPT-4.

The results of the regression analysis on QWK for each model are documented in **Table 2**. Similar to previous findings, significant recency bias and majority label bias were confirmed in the GPT-3.5 models. However, for the GPT-4 model, significant majority label bias was observed, but recency bias was not found. The impact of each bias varied between models. For instance, $L_l$ had a positive effect on QWK in GPT-3.5 (Jun23), whereas it had the opposite effect in GPT-3.5 (Nov23, Jan24). Moreover, the absolute values of significant regression coefficients were larger for GPT-3.5 than for GPT-4, indicating again that the influence of bias from examples on QWK was greater for GPT-3.5 than for GPT-4.

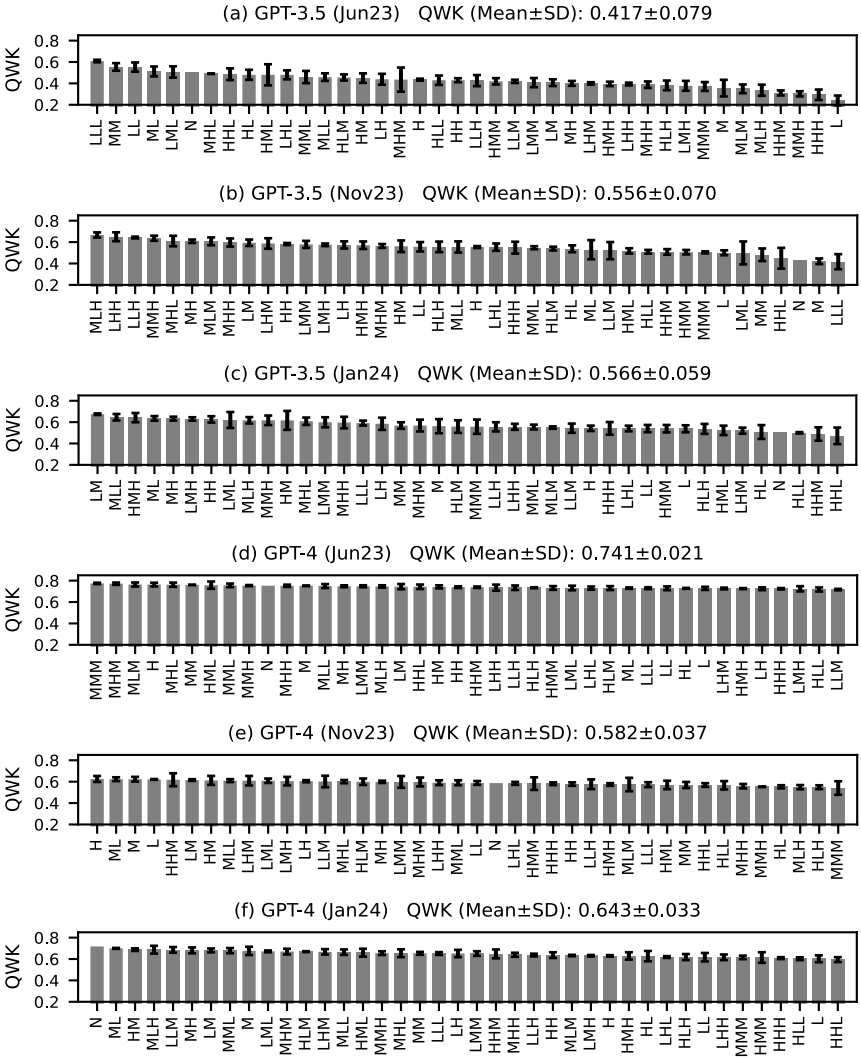

**Fig. 2.** QWK with variations of few-shot prompts across GPT models

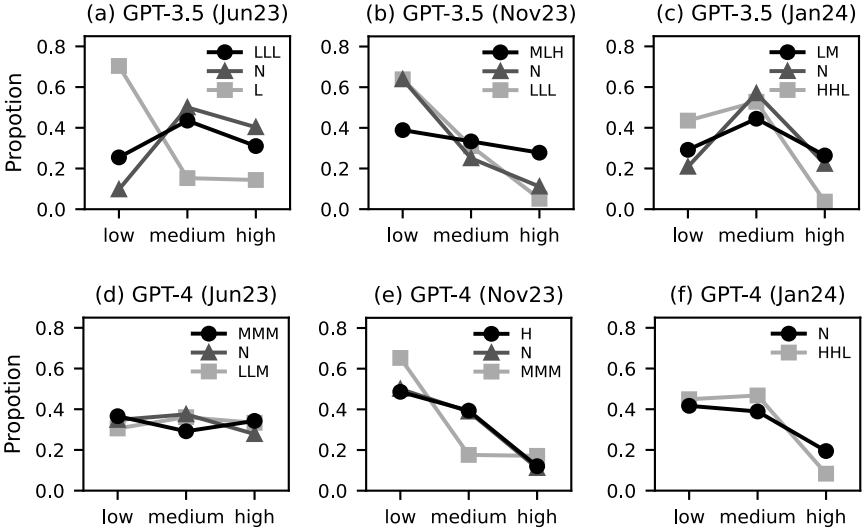

**Fig. 3.** Distribution of ratings in categories with the highest and lowest average QWK and zero-shot prompt across GPT models

**Table 1.** Average regression coefficients and coefficients of determination for essay scores (*: p<0.05, **: p<0.01, ***: p<0.001). Significant coefficients with the highest absolute value in the models are **bolded**. $\overline{\beta^s_{H_n}}, \overline{\beta^s_{M_n}}, \overline{\beta^s_{L_n}}, \overline{\beta^s_{H_l}}, \overline{\beta^s_{M_l}}, \overline{\beta^s_{L_l}}$ represent the average regression coefficients and $\overline{R^2}$ represents the average coefficient of determination.

| | Model | $\overline{\beta^s_{H_n}}$ | $\overline{\beta^s_{M_n}}$ | $\overline{\beta^s_{L_n}}$ | $\overline{\beta^s_{H_l}}$ | $\overline{\beta^s_{M_l}}$ | $\overline{\beta^s_{L_l}}$ | $\overline{R^2}$ |
|---|---|---|---|---|---|---|---|---|
| GPT-3.5 | Jun23 | -0.0024 | 0.0060* | -0.0036* | 0.0123*** | 0.0013 | **-0.0136***  | 0.2010 |
| | Nov23 | -0.0005 | 0.0021 | -0.0017 | 0.0136*** | 0.0037 | **-0.0173*** | 0.2094 |
| | Jan24 | -0.0049 | -0.0007 | 0.0057** | 0.0150*** | 0.0057** | **-0.0207*** | 0.2374 |
| GPT-4 | Jun23 | 0.0014 | -0.0013 | -0.0001 | 0.0020* | 0.0015 | **-0.0035** | 0.3040 |
| | Nov23 | 0.0023 | -0.0024 | 0.0001 | -0.0013 | **0.0023*** | -0.0010 | 0.1426 |
| | Jan24 | -0.0004 | -0.0005 | 0.0009 | -0.0013 | 0.0002 | 0.0011 | 0.2597 |

**Table 2.** Average regression coefficients and coefficients of determination for QWK (*: p<0.05, **: p<0.01, ***: p<0.001). Significant coefficients with the highest absolute value in the models are **bolded**.

| | Model | $\beta^q_{H_n}$ | $\beta^q_{M_n}$ | $\beta^q_{L_n}$ | $\beta^q_{H_l}$ | $\beta^q_{M_l}$ | $\beta^q_{L_l}$ | $R^2$ |
|---|---|---|---|---|---|---|---|---|
| GPT-3.5 | Jun23 | -0.0081 | -0.0089 | 0.0171* | -0.0194** | -0.0046 | **0.0240*** | 0.5165 |
| | Nov23 | -0.0108 | 0.0097 | 0.0012 | **0.0291*** | -0.0103 | -0.0189* | 0.3173 |
| | Jan24 | -0.0221*** | 0.0187** | 0.0034 | **0.0234*** | -0.0196** | -0.0038 | 0.3321 |
| GPT-4 | Jun23 | -0.0023 | **0.0088*** | -0.0064** | -0.0008 | -0.0034 | 0.0042 | 0.3144 |
| | Nov23 | 0.0014 | -0.0041 | 0.0027 | -0.0065 | 0.0065 | -0.0001 | 0.0764 |
| | Jan24 | **-0.0098*** | 0.0056 | 0.0043 | 0.0021 | 0.0001 | -0.0022 | 0.1665 |

## 5    Discussion

The impact of example selection on QWK in few-shot prompting varied across models. For instance, GPT-3.5 (Nov23, Jan24) performed better with most few-shot prompting conditions compared to zero-shot prompting (N), with some outperforming GPT-4 (Nov23). Notably, GPT-4 (Jan24) achieved the highest performance with zero-shot prompting. GPT-4 (Jun23) had the highest average QWK among all models and exhibited the lowest standard deviation in QWK, demonstrating robustness to example selection. Overall, comparing the standard deviation of QWK confirmed that GPT-3.5 models were more susceptible to the influence of example selection than GPT-4 models. Regarding practical implications, lower-tier models like GPT-3.5 were significantly influenced by examples, and careful example selection enabled them to outperform some higher-tier models like GPT-4. Therefore, when using lower-tier models for cost-effectiveness, it is recommended to carefully select examples. If cost is not a concern, using GPT-4 (Jun23), the most robust and highest-performing model, is advised. Although recent studies [33, 34] suggested that GPT-4's latest two models, Nov23 and Jan24, would have higher performance, Jun23 exhibited the best results. This demonstrates that the latest models do not always guarantee the highest performance, emphasizing the need for individual performance evaluations of each model, including minor versions. Furthermore, since performance varies across minor versions and previous research papers [7, 8] do not always include minor version information, research papers must include not only major version information but also minor version details when reporting model specifications and results.

Our results demonstrate the presence of biases, such as majority label bias and recency bias, in GPT-generated essay scores and QWK. These biases were more pronounced in GPT-3.5 compared to GPT-4. The impact on essay scores was more influenced by recency bias than majority label bias in GPT-3.5 models, while majority label bias was not significantly detected in GPT-4 models. The impact on QWK was more affected by recency bias in GPT-3.5 models, while GPT-4 only exhibited majority label bias. The impact of these biases varied across models, including minor versions. For example, presenting a lower-scoring essay last ($L_l$) had a positive effect on QWK in GPT-3.5 (Jun23) but the opposite effect in GPT-3.5 (Nov23, Jan24). The varying influence of examples on QWK across models could be attributed to differences in their inherent scoring tendencies. As depicted in Fig. 3, the evaluation tendencies with zero-shot prompts differed across models, including minor versions. For instance, GPT-3.5 (Jun23) tended to overestimate scores, while GPT-3.5 (Nov23) was inclined to underestimate them. Consequently, $L_l$ could improve QWK for GPT-3.5 (Jun23) by moderating potentially higher scores but decrease QWK for GPT-3.5 (Nov23) by further lowering scores. However, this explanation does not apply universally to other phenomena, indicating a need for further verification.

The coefficient of determination for QWK in GPT-3.5 (Jun23) exceeded 0.5, but all other coefficients of determination were below 0.4. This suggests that factors other than the ratings of examples are influencing the essay scores and QWK. This research differentiated examples solely based on their scores, thus not conducting an analysis focused on individual examples. Future investigations could enhance our understanding of the impact examples have by incorporating analysis of information beyond scores, such as linguistic characteristics inherent in each example, suggesting a potential for deeper insights into how examples influence outcomes.

Due to the necessity of conducting experiments with multiple models under diverse conditions, we employed a method to extract as diverse a set of essays as possible, albeit with a limited evaluation essay dataset. Nevertheless, the ability to compare models under these conditions yielded valuable insights explained above. Future research can focus on obtaining more detailed findings by increasing the sample size of evaluation essays. Furthermore, while this study focused on GPT models, future research could explore other LLMs like Gemini, Claude, LLaMA, and Vicuna for cross-model insights. Additionally, using different datasets such as the Automated Student Assessment Prize (ASAP) program data [35] or The Cambridge Learner Corpus-First Certificate in English exam (CLC-FCE) [36] could reveal how dataset choice impacts results.

## 6  Conclusion

In this study, we have explored the impact of example selection on the performance of AES using few-shot prompting with various GPT models. Our findings reveal that the choice and order of examples in few-shot prompts influence the agreement between GPT and human rater scores, as measured by QWK. The impact of example selection varies across models, with GPT-3.5 being more sensitive to the selection of examples compared to GPT-4. Through regression analysis, we have identified the presence of majority label bias and recency bias in GPT-generated essay scores and QWK. These biases are more evident in GPT-3.5 models, while GPT-4 exhibits greater robustness. Interestingly, we find that careful example selection can enable GPT-3.5 models to outperform some GPT-4 models, highlighting the importance of thoughtful prompt engineering. Our study underscores the need for individual performance evaluations of each model, including minor versions, as the impact of example selection varies among models and the latest models do not always guarantee the highest performance.

**Acknowledgment.** In preparing this manuscript, I used DeepL, Grammarly, ChatGPT, and Claude to improve the language. The tools did not contribute to generating any original ideas. This work was supported by JSPS KAKENHI Grant Number 23K02707 and the research program on "Creation of generative AI learning environment for teachers," conducted at the Tokyo Foundation for Policy Research.

This preprint has not undergone any post-submission improvements or corrections. The Version of Record of this contribution is published in Communications in Computer and Information Science, vol 2150, and is available online at https://doi.org/10.1007/978-3-031-64315-6_5.